\documentclass[11pt]{article}
\usepackage{fullpage,url,natbib}
\usepackage{amsthm,amsfonts,amsmath,amssymb,epsfig,color,float,graphicx,verbatim}
\usepackage{algorithm}
\usepackage{algorithmic}

\usepackage{enumitem}
\usepackage{hyperref}
\hypersetup{
	colorlinks   = true, 
	urlcolor     = blue, 
	linkcolor    = blue, 
	citecolor   = blue 
}

\newtheorem{theorem}{Theorem}
\newtheorem{proposition}{Proposition}
\newtheorem*{proposition*}{Proposition}

\newtheorem{remark}{Remark}

\renewcommand{\eqref}[1]{Eq.~(\ref{#1})}

\newcommand{\thmref}[1]{Theorem~\ref{#1}}

\newcommand{\reals}{\mathbb{R}}

\newcommand{\N}{\mathbb{N}}

\newcommand{\norm}[1]{\Vert #1 \Vert}

\newcommand{\vertiii}[1]{{\left\vert\kern-0.25ex\left\vert\kern-0.25ex\left\vert #1 
    \right\vert\kern-0.25ex\right\vert\kern-0.25ex\right\vert}}

\newcommand{\mexp}{\mathbb{E}}

\newcommand{\E}{\mathop{\mexp}}

\newcommand{\eps}{\varepsilon}

\usepackage{xcolor}

\newcommand{\beq}{\begin{eqnarray*}}
\newcommand{\eeq}{\end{eqnarray*}}
\newcommand{\beqn}{\begin{eqnarray}}
\newcommand{\eeqn}{\end{eqnarray}}

\usepackage{ifmtarg}
\usepackage{xifthen}%
\newcommand{\ent}[1][]{%
\ifthenelse{\isempty{#1}}{%
\mathrm{H}
}{
\mathrm{H}^{(#1)}
}}

\newcommand{\loch}[1][]{%
\ifthenelse{\isempty{#1}}{%
\mathrm{h}
}{
\mathrm{h}^{(#1)}
}}

\newcommand{\hide}[1]{}

\newcommand{\Acal}{\mathcal{A}}
\newcommand{\Bcal}{\mathcal{B}}
\newcommand{\Dcal}{\mathcal{D}}
\newcommand{\Fcal}{\mathcal{F}}

\newcommand{\Lcal}{\mathcal{L}}
\newcommand{\Ncal}{\mathcal{N}}

\newcommand{\Nbr}{\mathcal{N}_{[\,]}}

\newcommand{\NN}{\mathbb{N}}
\newcommand{\Lip}{\mathrm{Lip}}
\newcommand{\Hol}{\mathrm{H\ddot{o}l}}

\newtheorem*{rep@theorem}{\rep@title}
\newcommand{\newreptheorem}[2]{%
\newenvironment{rep#1}[1]{%
 \def\rep@title{#2 \ref{##1}}%
 \begin{rep@theorem}}%
 {\end{rep@theorem}}}
\makeatother

\newreptheorem{proposition}{Proposition}

\title{\bf{Efficient Agnostic Learning with Average Smoothness}
}

\author{
 Steve Hanneke\thanks{\texttt{steve.hanneke@gmail.com}}
 \vspace{.05in}
 \\Purdue University
 \and Aryeh Kontorovich\thanks{\texttt{karyeh@cs.bgu.ac.il}
 }
 \vspace{.05in}
 \\Ben-Gurion University\\of the Negev
 \and Guy Kornowski\thanks{\texttt{guy.kornowski@weizmann.ac.il}}  \vspace{.05in}
 \\Weizmann Institute\\of Science
}

\date{}

\begin{document}

\maketitle
\begin{center}\vspace{-1cm}\today\vspace{0.5cm}\end{center}

\begin{abstract}
We study distribution-free nonparametric regression following a notion of average smoothness initiated by \citet{ashlagi2021functions}, which measures the ``effective'' smoothness of a function with respect to an arbitrary unknown underlying distribution. 
While the recent work of \citet{hanneke2023near} 
established tight uniform convergence bounds for average-smooth functions in the realizable case and provided a computationally efficient realizable learning algorithm, both of these results currently lack analogs in the general agnostic (i.e. noisy) case.

In this work, we fully close these gaps. First, we provide a distribution-free uniform convergence bound for average-smoothness classes in the agnostic setting. Second, we match the derived sample complexity with a computationally efficient agnostic learning algorithm. Our results, which are stated in terms of the intrinsic geometry of the data and hold over any totally bounded metric space, show that the guarantees recently obtained for realizable learning of average-smooth functions transfer to the agnostic setting.
At the heart of our proof, we establish the uniform convergence rate of a function class in terms of its bracketing entropy, which may be of independent interest.
\end{abstract}

\section{Introduction}

Numerous frameworks in learning theory and statistics 
formalize
the 
intuitive
insight
that ``smooth functions are easier to learn than rough ones''
\citep{MR1920390, tsybakov2008nonparametric,gine2021mathematical}.
The various
measures of smoothness that were studied in a statistical context
include
the popular
Lipschitz or H\"older seminorms; the bounded variation norm \citep{DBLP:journals/iandc/Long04}; Sobolev, Sobolev-Slobodetskii and Besov norms \citep{MR2324525}; averaged modulus of continuity \citep{MR995672,MR2761605}; and probabilistic Lipschitzness in the context of classification \citep{urner:13a,urner:13,DBLP:conf/colt/KpotufeUB15}.

\begin{figure}
\begin{center}
	\includegraphics[trim=0cm 4cm 0cm 3cm,clip=true, width=0.9\textwidth]{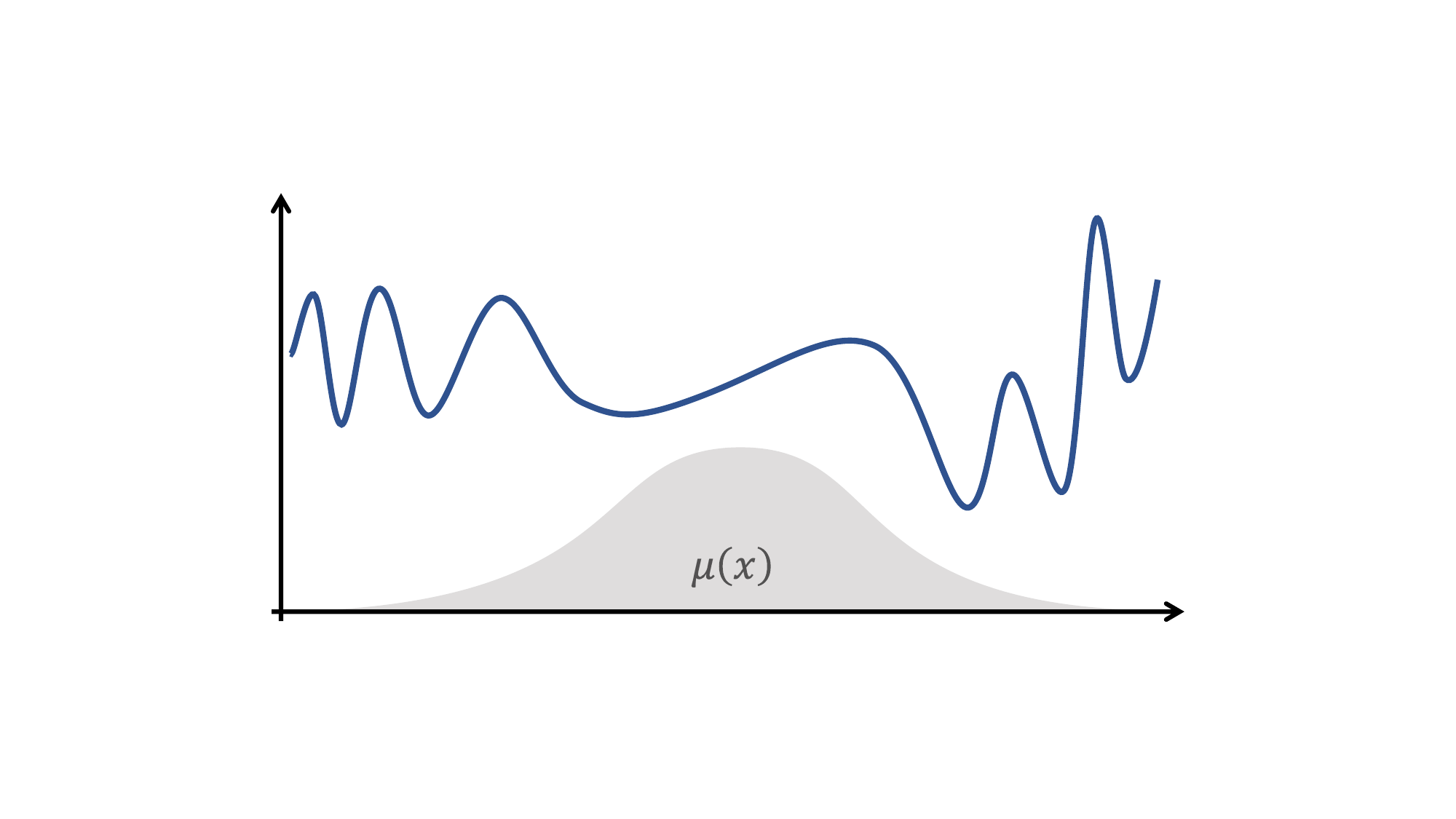}%
	\caption{Illustration of a function and a measure $\mu$ exhibiting a large gap between ``worst-case'' smoothness (occurring in low density regions) and average-smoothness with respect to $\mu$; figure taken from \citep{hanneke2023near}.}
	\label{fig:avg_smooth}
	\end{center}
\end{figure}

In particular, a recent line of work \citep{ashlagi2021functions, hanneke2023near} studied a notion of \emph{average smoothness}
with respect to an arbitrary measure. Informally, the average smoothness is defined by considering the ``local'' H\"older (or Lipschitz) smoothness of a function at each point of the instance space, averaged with respect to the marginal distribution over the space; see Figure \ref{fig:avg_smooth} for a simple illustration, and Section~\ref{subsec: avg smooth} for a formal definition.
The main conclusion of the aforementioned works is that it is possible to guarantee statistical generalization solely in terms of the average smoothness
for any underlying measure, effectively replacing the classic H\"older (or Lipschitz) constant with a much tighter distribution-dependent quantity. In particular, \citet{hanneke2023near} proved a uniform convergence bound for the class of on-average-smooth functions in the realizable (i.e. noiseless) case, and complemented this result with an efficient realizable learning algorithm. 
With regard to the 
the general case of agnostic learning, the results of \citeauthor{hanneke2023near} had some limitations.
In particular,
the 
general reduction from agnostic to realizable learning
\citep{hopkins2022realizable}
deployed therein
left two unfulfilled desiderata. 
From a statistical perspective, 
it remained
open whether
a function class with bounded average smoothness under some distribution $\mu$ is $\mu$-Glivenko-Cantelli, namely that the excess risk decays uniformly over the class;
only the existence of 
{\em some} returned
predictor with
small excess risk was established.
On the computational side, the agnostic algorithm is highly inefficient:
its runtime complexity is exponential in the sample size,
in contrast with the polynomial-time realizable algorithm.

\subsection{Our Contributions.}

In this paper we study distribution-free agnostic learning of average-smooth functions, and address the issues raised above. Our main contributions can be summarized as follows:

\begin{itemize}
    \item \textbf{Agnostic uniform convergence (Theorem~\ref{thm: bracket to uc} and Theorem~\ref{thm: gen}).}
    We provide a distribution-free uniform convergence bound for the class of average-smooth functions in the agnostic setting (Theorem~\ref{thm: gen}). 
This bound actually follows from   
a more general result, in which we bound the uniform convergence in terms of the bracketing entropy of the class (Theorem~\ref{thm: bracket to uc}).
The latter is
widely
applicable and may be of independent interest.

    \item \textbf{Efficient agnostic algorithm (Theorem~\ref{thm: alg}).}
    We present a polynomial time algorithm for agnostic learning of on-average-smooth functions. 
The resulting    
    sample complexity matches the aforementioned uniform convergence bound, which also matches that of the exponential-time agnostic algorithm of \citet{hanneke2023near}. Furthermore, the algorithm's running time matches that of their efficient realizable learning algorithm.

\end{itemize}

\section{Preliminaries}

\paragraph{Setting.}
Throughout the paper we consider functions $f:\Omega\to[0,1]$ where $(\Omega,\rho)$ is a metric space. We will consider a distribution $\Dcal$ over $\Omega\times[0,1]$ with marginal $\mu$ over $\Omega$, such that $(\Omega,\rho,\mu)$ forms a metric probability space (namely, $\mu$ is supported on the Borel
$\sigma$-algebra induced by $\rho$).
For any measurable function $f:\Omega\to[0,1]$ we associate its $L_1$ risk ${ 
L_\Dcal(f):=  \E_{ (X,Y)\sim\Dcal}}|f(X)-Y|$, and its empirical risk with respect to a sample $S=(X_i,Y_i)_{i=1}^{n}\sim\Dcal^n:~L_S(f):=\frac{1}{n}\sum_{i=1}^{n}|f(X_i)-Y_i|$.
More generally, we associate to any measurable function its $L_1$ norm $
{
\norm{f}_{L_1(\mu)}:=\E_{X\sim\mu}|f(X)|
}
$, and given a sample $(X_1,\dots,X_n)\sim\mu^n$, we denote its $L_1$ norm with respect to the empirical measure by $\norm{f}_{L_1(\mu_n)}:=\frac{1}{n}\sum_{i=1}^{n}|f(X_i)|$.

\paragraph{Metric notions.}
We denote by $B(x,r):=\{x'\in\Omega:\rho(x,x')\leq r\}$ the closed ball around $x\in\Omega$ of radius $r>0$.
For $t>0,~A,B\subset\Omega$, we say that $A$ is a $t$-\emph{cover} of $B$ if $B\subset\bigcup_{a\in A}B(a,t)$, and define the $t$-\emph{covering number} $\Ncal_B(t)$ to be the minimal cardinality of any $t$-cover of $B$.
We say that $A\subset B\subset \Omega$ is a $t$-\emph{packing} of $B$ if $\rho(a,a')\geq t$ for all $a\neq a'\in A$. We call $V$ a $t$-\emph{net} of $B$ if it is a $t$-cover and a $t$-packing.
A metric space $(\Omega,\rho)$ is said to be \emph{doubling} 
with constant $D\in\N$
if every
ball $B\subset\Omega$ of radius $r$ 
verifies
$
\Ncal_B(r/2)\le D
$.
The \emph{doubling dimension} is defined as $\inf_{D\in\N}\log_2 D$, where the 
infimum runs over all valid doubling constants for $(\Omega,\rho)$.

\paragraph{Bracketing.}

Given any two functions $l,u:\Omega\to[0,1]$, we say that $f:\Omega\to[0,1]$ belongs to the \emph{bracket} $[l,u]$ if $l\leq f\leq u$. A set of brackets $\Bcal$ is said to cover a function class $\Fcal$ if every function in $\Fcal$ belongs to some bracket in $\Bcal$.
We say that $[l,u]$ is a $t$-bracket with respect to a norm $\|\cdot\|$ if $\norm{u-l}\leq t$. The $t$-\emph{bracketing number} $\Nbr(\Fcal,\|\cdot\|,t)$ is defined as the minimal cardinality of any set of $t$-brackets that covers $\Fcal$. The logarithm of this quantity is called the \emph{bracketing entropy}.

\begin{remark}[Covering vs. bracketing] 
Having recalled two notions that quantify the ``size'' of a normed function space $(\Fcal,\|\cdot\|)$
--- namely, its 
covering 
and 
bracketing numbers
--- it is useful to note they are related through $\Ncal_{\Fcal}(\eps)\leq \Nbr(\Fcal,\|\cdot\|,2\eps)$,
though no converse inequality of this sort holds in general. 
On the other hand, the main advantage of using bracketing numbers for generalization bounds is that it suffices to bound the \textbf{ambient} bracketing numbers with respect to the distribution-specific metric, as opposed to the \textbf{empirical} covering numbers which are necessary to guarantee generalization \citep[Section 2.1.1]{van1996weak}.
\end{remark}

\subsection{Average smoothness \citep{ashlagi2021functions,hanneke2023near}.} \label{subsec: avg smooth}

The definition of average smoothness closely follows that given by \citet{hanneke2023near}.
For $\beta\in(0,1]$ and $f:\Omega\to\reals$, we define its $\beta$-slope at $x\in\Omega$ to be
\[
\Lambda_f^\beta(x):=\sup_{y\in\Omega\setminus\{x\}}\frac{|f(x)-f(y)|}{\rho(x,y)^\beta}~.
\]
Recall that $f$ is called $\beta$-H\"older continuous if 
\[
\norm{f}_{\Hol^\beta}:=\sup_{x\in\Omega}\Lambda_f^\beta(x)<\infty~;
\]
the latter is known as the
H\"older seminorm. In particular, when $\beta=1$, these are exactly the Lipschitz functions equipped with the Lipschitz seminorm.
For a metric probability space $(\Omega,\rho,\mu)$, we consider the \emph{average} $\beta$-slope to be the mean of $\Lambda_f^\beta(X)$ where $X\sim\mu$. We define
\begin{align*}
\overline{\Lambda}_f^\beta(\mu)&:=\E_{X\sim\mu}\left[\Lambda_f^\beta(X)\right]
~.
\end{align*}
Notably
\begin{equation} \label{eq:Lambda ineq}
\overline{\Lambda}_f^\beta(\mu)
\leq \norm{f}_{\Hol^\beta}~,
\end{equation}
where the gap can be \emph{infinitely} large, as demonstrated by \citet{hanneke2023near}.
The notion of average smoothness induces
the corresponding function class (alongside the classic ``worst-case'' one):
\begin{align*}
    \Hol^\beta_{H}(\Omega)&:=\left\{f:\Omega\to[0,1]\,:\,\norm{f}_{\Hol^\beta} \leq H\right\}~,
    \\
    \overline{\Hol}^\beta_{H}(\Omega,\mu)&:=\left\{f:\Omega\to[0,1]\,:\,\overline{\Lambda}_f^\beta(\mu)\leq H\right\}~.
\end{align*}
We occasionally omit $\mu$ when it is clear from context.  Note that for any measure $\mu:$
\[
\Hol^\beta_{H}(\Omega)\subset\overline{\Hol}^\beta_{H}(\Omega,\mu)
\]
due to \eqref{eq:Lambda ineq}, where the containment is strict in general. 
The special case of $\beta=1$ recovers the 
average-Lipschitz
function class $\Lip_H(\Omega)\subset\overline{\Lip}_H(\Omega,\mu)
$ studied by \citet{ashlagi2021functions}, while the general case $\beta\in(0,1]$ was studied by \citet{hanneke2023near}.

In particular, we will now recall one of the main results of \citet{hanneke2023near} which establishes a bound on the bracketing entropy of average-smoothness classes. Crucially, the bound does not depend on $\mu$, which allows to obtain distribution-free generalization guarantees.

\begin{theorem}[\citealp{hanneke2023near}, Theorem 1] \label{thm: bracket}
For any metric probability space $(\Omega,\rho,\mu)$, any $\beta\in(0,1]$ and any $0<\eps<H:$
\begin{align*}
\log\Nbr(\overline{\Hol}^\beta_H(\Omega,\mu),L_1(\mu),\eps)
&\leq
\Ncal_{\Omega}\left(\left(\frac{\eps}{128H\log(1/\eps)}\right)^{1/\beta}\right)\cdot\log\left(\frac{16\log_2(1/\eps)}{\eps}\right)
~.
\end{align*}
\end{theorem}

\begin{remark}[Weak average]
    \cite{hanneke2023near} also considered the even larger space of functions which are weakly-average-smooth, namely such that $\sup_{t>0}\,t \cdot\mu(x:\Lambda_f^\beta(x)\geq t)\leq H$. Note that this class is indeed larger than $\overline{\Hol}^\beta_{H}(\Omega,\mu)$ by Markov's inequality. The bracket entropy bound in Theorem~\ref{thm: bracket} was actually proven for this even larger class. Consequently, all the uniform convergence results to appear in the next section also hold for this larger class. We choose to focus on the class $\overline{\Hol}^\beta_{H}(\Omega,\mu)$ throughout this paper for ease of presentation.
\end{remark}

\section{Generalization bounds}

Our first goal is to establish a uniform convergence result for the class $\overline{\Hol}^\beta_H(\Omega,\mu)$, which holds regardless of the distribution $\Dcal$ whose marginal is $\mu$ (in particular, the bound does not depend on $\mu$).
Notably, a bound of this sort was previously established by \citet{hanneke2023near} only for $\Dcal$ that are realizable by the function class, namely for which there exists an $f^*\in\overline{\Hol}^\beta_H(\Omega,\mu)$ with $L_\Dcal(f^*)=0$.

In order to leverage Theorem~\ref{thm: bracket} towards establishing an agnostic risk bound, we prove what is apparently a novel uniform deviation bound in terms
of bracketing numbers:

\begin{theorem} \label{thm: bracket to uc}
Suppose $(\Omega,\rho)$ is a metric space, $\Fcal\subseteq [0,1]^\Omega$ is a function class, and let $\Dcal$ be a distribution over $\Omega\times[0,1]$ with marginal $\mu$ over $\Omega$. Then with probability at least $1-\delta$ over drawing a sample $S\sim\Dcal^n$ it holds that for all $f\in\Fcal,~\alpha\geq0:$
\[
\left|L_{\Dcal}(f)-L_S(f)\right|\leq \alpha+O\left(\sqrt{\frac{\log \Nbr(\Fcal,L_1(\mu),\alpha)+\log(1/\delta)}{n}}\right)~.
\]
\end{theorem}

\begin{remark}[Other losses]
The proof of Theorem~\ref{thm: bracket to uc} is the only place throughout the paper that relies on the considered risk being with respect to the $L_1$ loss. In particular, in Eq.~(\ref{eq: loss bracket}) we prove an analog of the contraction lemma (cf. \citealp[Lemma 5.7]{mohri2018foundations}) for bracketing entropies with respect to the $L_1$ loss. This statement holds with essentially the same proof
under mild assumption on the loss, e.g. as long as the loss $\ell(f(x),y)$ is symmetric with respect to exchanging its variables, monotone and Lipschitz with respect to $|f(x)-y|$ (with an incurred dependence on the Lipschitz constant). In particular, since the functions discussed in this paper are bounded, the results are readily extendable to $L_p$ losses for any $p\in[1,\infty)$ (naturally yielding $p$-dependent rates due to the $p$-dependent Lipschitz constant).
\end{remark}

In our case of interest, plugging the bracket entropy  bound for average smoothness classes from Theorem~\ref{thm: bracket} 
into the uniform deviation bound in Theorem~\ref{thm: bracket to uc}
yields the following:

\begin{theorem} \label{thm: gen}
For any metric space $(\Omega,\rho)$ and distribution $\Dcal$ with marginal $\mu$ as above, it holds with probability at least $1-\delta$ over drawing a sample $S\sim\Dcal^n$ that for all $f\in\overline{\Hol}_{H}^\beta(\Omega,\mu),~\alpha\geq0:$
    \[
    \left|L_{\Dcal}(f)-L_S(f)\right|
    =\alpha+\widetilde{O}\left(\sqrt{\frac{\Ncal_{\Omega}\left(\left(\frac{\alpha}{128H\log(1/\alpha)}\right)^{1/\beta}\right)+\log(1/\delta)}{n}}\right)
    ~.
    \]
\end{theorem}

\begin{remark}[Doubling metrics]
In most cases of interest, $(\Omega,\rho)$ is a doubling metric space of some dimension $d$,\footnote{Namely, any ball of radius $r>0$ can be covered by $2^d$ balls of radius $r/2$.} e.g. when $\Omega$ is a subset of $\reals^d$ (or more generally a $d$-dimensional Banach space).
For $d$-dimensional doubling spaces of finite diameter we have
$\Ncal_{\Omega}(\eps)\lesssim\left(\frac{1}{\eps}\right)^d$ \citep[Lemma 2.1]{gottlieb2016adaptive}, which by plugging into Theorem~\ref{thm: gen} and optimizing over $\alpha\geq0$ yields the simplified generalization bound
\[
\sup_{f\in\overline{\Hol}_{H}^\beta(\Omega,\mu)}\left|L_{\Dcal}(f)-
L_\Dcal(f)
\right|
=\widetilde{O}\left(\frac{H^{d/(d+2\beta)}}{n^{\beta/(d+2\beta)}}\right)~.
\]
Equivalently, $\sup_{f\in\overline{\Hol}_{H}^\beta(\Omega,\mu)}\left|L_{\Dcal}(f)-L_S(f)\right|\leq\eps$ whenever $n\geq N$ for
\[
N=\widetilde{O}\left(\frac{H^{d/\beta}}{\eps^{(d+2\beta)/\beta}}\right)
~,
\]
up to a constant that depends
(exponentially)
on $d$, but is independent of $H,\eps$.
\end{remark}

\section{Efficient agnostic learning algorithm}

Having established the sample complexity required for controlling the excess risk uniformly over average-smooth functions, we turn to seeking an efficient agnostic regression algorithm that attains this sample complexity. We note that this is a nontrivial task due to the nature of the class $\overline{\Hol}_{H}^\beta(\Omega,\mu)$, which is \emph{unknown to the learner}. Indeed, without knowledge of the underlying distribution, the learner cannot evaluate 
a candidate function's average smoothness with respect to the given distribution,
thus a naive empirical-risk-minimization approach over the function class is inapplicable.
The key to designing an average-smooth regression algorithm is the analysis of the \emph{empirical smoothness} induced by the sample, namely the quantity
\[
\widehat{\Lambda}^\beta_f:=\frac{1}{n}\sum_{i=1}^{n}\max_{X_j\neq X_i}\frac{|f(X_i)-f(X_j)|}{\rho(X_i,X_j)^\beta}~
\]
for any function $f:\Omega\to[0,1]$. \citet{hanneke2023near} proved a tail bound for the empirical smoothness in terms of the true average smoothness. Subsequently, their agnostic algorithm is a certain exhaustive search procedure over the space of empirically-smooth functions, and thus highly inefficient. In particular, the runtime of the algorithm is exponential in the sample size, 
which provided ample motivation to seek an efficient one.

In the following theorem we provide a polynomial-time algorithm that matches the same sample complexity, 
thus closing the exponential gap.

\begin{theorem} \label{thm: alg}
There is a polynomial time algorithm $\Acal$ such that for any metric space $(\Omega,\rho)$,
any $\beta\in(0,1],~0<\eps<H$, and any distribution $\Dcal$ over $\Omega\times[0,1]$, given a sample $S\sim\Dcal^n$ of size $n\geq N$ where
$N=N(\beta,\eps,\delta)$
satisfies
\[
N=\widetilde{O}\left(
\frac{\Ncal_{\Omega}\left(\left(\frac{\eps}{640H\log(1/\eps)}\right)^{1/\beta}\right)+\log(1/\delta)}
{\eps^2}
\right),
\]
the algorithm constructs a hypothesis $f=\Acal(S)$ such that 
\[
L_\Dcal(f)\leq \inf_{f^*\in\overline{\Hol}_{H}^\beta(\Omega,\mu)}L_\Dcal(f^*)+\eps
\]
with probability at least
$1-\delta$.
\end{theorem}

\begin{remark}[Doubling metrics]
As previously discussed, in most cases of interest we have
$\Ncal_{\Omega}(\eps)\lesssim\left(\frac{1}{\eps}\right)^d$ for some dimension $d\in\NN$. 
That being the case, Theorem~\ref{thm: alg} yields the simplified sample complexity bound
\[
N=\widetilde{O}\left(\frac{H^{d/\beta}}{\eps^{(d+2\beta)/\beta}}\right)~,
\]
or equivalently
\[
L_{\Dcal}(f)=
\inf_{f^*\in\overline{\Hol}_{H}^\beta(\Omega,\mu)}L_\Dcal(f^*)
+\widetilde{O}\left(\frac{H^{d/(d+2\beta)}}{n^{\beta/(d+2\beta)}}\right)~,
\]
up to a constant that depends
(exponentially)
on $d$, but is independent of $H,n$.
\end{remark}

\begin{remark}[Computational complexity]
The algorithm 
described in \thmref{thm: alg}
involves a single preprocessing step with runtime $\widetilde{O}(n^{2\omega})$ where $\omega\approx2.37$ is the current matrix multiplication exponent,
after which
$f(x)$
can be evaluated at any
given
$x\in\Omega$
in $O(n^2)$ time.
We note that the computation at inference time matches that of (classic) Lipschitz/H\"older regression (e.g. \citealp{7944658}).
\end{remark}

We will now outline the proof of Theorem~\ref{thm: alg}, which appears in Section~\ref{sec: alg proof} along the full description of the algorithm.
Denoting the Bayes-optimal risk by
$L^*=\inf_{f^*\in\overline{\Hol}_{H}^\beta(\Omega,\mu)}L_\Dcal(f^*)$, we assume
without loss of generality (by a standard approximation argument) that the infimum is obtained, and let $f^*\in\overline{\Hol}_{H}^\beta(\Omega,\mu)$ be a function with $L_\Dcal(f^*)=L^*$.
Given a sample $(X_i,Y_i)_{i=1}^{n}\sim\Dcal^n$, the algorithm first constructs labels $(\widehat{f}(X_i))_{i=1}^{n}$ such that 
\begin{equation} \label{eq: proof sketch 1}
    L_S(\widehat{f})\leq L^*+\frac{\eps}{3}
\end{equation}
and
\begin{equation} \label{eq: proof sketch 2}
\widehat{\Lambda}^\beta_{\widehat{f}}\lesssim H~.
\end{equation}
We show that such 
a ``relabeling'' is
obtainable by solving a linear program which minimizes the empirical error under the empirical smoothness constraint.
This program is feasible since $f^*$ satisfies both conditions with high probability. Indeed, $f^*$ satisfies \eqref{eq: proof sketch 1} by Theorem~\ref{thm: gen} (for large enough sample size), while \eqref{eq: proof sketch 2} follows from the aforementioned tail bound of empirical smoothness.
With these labels in hand, we invoke an approximate-extension procedure due to \citet{hanneke2023near} that extends $\widehat{f}$ to $f:\Omega\to[0,1]$ satisfying $L_S(f)\leq L_S(\widehat{f})+\frac{\eps}{3}$ and $\overline{\Lambda}^\beta_f(\mu)\lesssim \widehat{\Lambda}^\beta_{\widehat{f}}$ 
with high probability.
Combining the latter property with \eqref{eq: proof sketch 2} yields $\overline{\Lambda}^\beta_f(\mu)\lesssim H$. Thus, we have overall obtained some $f$ in the average-smooth class (with a slightly inflated average-smoothness parameter) whose empirical risk is bounded according to \eqref{eq: proof sketch 1} by
\[
L_S(f)\leq L_S(\widehat{f})+\frac{\eps}{3}
\leq L^*+\frac{2\eps}{3}~.
\]
Finally, invoking Theorem~\ref{thm: gen} we conclude that the smooth-on-average $f$ has small excess risk, resulting in
\[
L_\Dcal(f)
\leq L_S(f)+\frac{\eps}{3}
\leq L^*+\eps
\]
with high probability, whenever the sample is large enough.

\section{Proofs}

\subsection{Proof of Theorem~\ref{thm: bracket to uc}}

We start by denoting the loss class $\Lcal_{\Fcal}\subseteq [0,1]^{\Omega\times[0,1]}:$
\begin{equation} \label{eq: loss class}
\Lcal_\Fcal:=\left\{\ell_f(x,y):=|f(x)-y|:f\in\Fcal\right\}~.
\end{equation}
We will show that for any $\alpha>0$, the bracketing entropy of $\Lcal_\Fcal$ is no larger than that of $\Fcal$, namely
\begin{equation} \label{eq: loss bracket}
\Nbr(\Lcal_{\Fcal},L_1(\mu),\alpha)\leq\Nbr(\Fcal,L_1(\mu),\alpha)~.
\end{equation}
To that end, fix $\alpha>0$, let $\Bcal_\alpha$ be a minimal $\alpha$-bracket of $\Fcal$, and denote for any $f\in\Fcal$ 
by $[f_L,f_U]\in\Bcal_\alpha$ its associated bracket. For $\ell_f\in\Lcal_\Fcal$ as defined in \eqref{eq: loss class},
we define the bracket $[(\ell_f)_L,(\ell_f)_U]$ as
\[
\left((\ell_f)_L(x,y),(\ell_f)_U)(x,y)\right)
:=\begin{cases}
    (0,f_U(x)-f_L(x)), & \mathrm{if~~}f_L(x)\leq y\leq f_U(x)
    \\
    (f_L(x)-y,f_U(x)-y), & \mathrm{if~~}y<f_L(x)
    \\
    (y-f_U(x),y-f_L(x)), & \mathrm{if~~}y>f_U(x)
    ~.
\end{cases}
\]
Notice that this is indeed a valid bracket, since  $f_L(x)\leq f(x)\leq f_U(x)$
implies that
for any $(x,y)\in\Omega\times[0,1]:$
\begin{align*}
(\ell_f)_L(x,y)
&=
\begin{cases}
    0, & \mathrm{if~~}f_L(x)\leq y\leq f_U(x)
    \\
    f_L(x)-y, & \mathrm{if~~}y<f_L(x)
    \\
    y-f_U(x), & \mathrm{if~~}y>f_U(x)
\end{cases}
\\
&\leq
\begin{cases}
    |f(x)-y|, & \mathrm{if~~}f_L(x)\leq y\leq f_U(x)
    \\
    f(x)-y, & \mathrm{if~~}y<f_L(x)
    \\
    y-f(x), & \mathrm{if~~}y>f_U(x)
\end{cases}
\\
&=
|f(x)-y|
=\ell_f(x,y)
~,
\end{align*}
and similarly
\begin{align*}
(\ell_f)_U(x,y)
&=
\begin{cases}
    f_U(x)-f_L(x), & \mathrm{if~~}f_L(x)\leq y\leq f_U(x)
    \\
    f_U(x)-y, & \mathrm{if~~}y<f_L(x)
    \\
    y-f_L(x), & \mathrm{if~~}y>f_U(x)
\end{cases}
\\
&\geq
\begin{cases}
    |f(x)-y|, & \mathrm{if~~}f_L(x)\leq y\leq f_U(x)
    \\
    f(x)-y, & \mathrm{if~~}y<f_L(x)
    \\
    y-f(x), & \mathrm{if~~}y>f_U(x)
\end{cases}
\\
&=
|f(x)-y|=\ell_f(x,y)
~.
\end{align*}
Moreover, by construction we see that for any $(x,y)\in\Omega\times[0,1]:(\ell_f)_U(x,y)-(\ell_f)_L(x,y)=f_U(x)-f_L(x)$, hence 
\[
\norm{(\ell_f)_U-(\ell_f)_L}_{L_1(\Dcal)}
\leq \norm{f_U-f_L}_{L_1(\mu)}\leq \alpha~,
\]
showing we indeed constructed an $\alpha$-bracket. As it is clearly of size at most $|\Bcal_\alpha|$, we proved \eqref{eq: loss bracket}.

Now note that for any $f\in\Fcal:$
\begin{align*}
L_\Dcal(f)-L_S(f)
&=\norm{\ell_f}_{L_1(\Dcal)}-\norm{\ell_f}_{L_1(\Dcal_n)}
\\
&\leq 
\norm{\ell_f-(\ell_f)_L}_{L_1(\Dcal)}
+\norm{(\ell_f)_L}_{L_1(\Dcal)}
-\norm{\ell_f}_{L_1(\Dcal_n)}
\\
&\leq 
\norm{(\ell_f)_U-(\ell_f)_L}_{L_1(\Dcal)}
+\norm{(\ell_f)_L}_{L_1(\Dcal)}
-\norm{\ell_f}_{L_1(\Dcal_n)}
\\
&\leq
\alpha + \norm{(\ell_f)_L}_{L_1(\Dcal)} -\norm{\ell _f}_{L_1(\Dcal_n)}
\\
&\leq
\alpha + \norm{(\ell_f)_L}_{L_1(\Dcal)} -\norm{(\ell _f)_L}_{L_1(\Dcal_n)}~,
\end{align*}
hence
\[
\sup_{f\in\Fcal}\left(L_\Dcal(f)-L_S(f)\right)
\leq \alpha + \max_{(\ell_f)_L}\left(\norm{(\ell_f)_L}_{L_1(\Dcal)} -\norm{(\ell _f)_L}_{L_1(\Dcal_n)}\right)~.
\]
Similarly, we also have
\[
\sup_{f\in\Fcal}\left(L_S(f)-L_\Dcal(f)\right)
\leq \alpha + \max_{(\ell_f)_U}\left(\norm{(\ell_f)_U}_{L_1(\Dcal_n)} -\norm{(\ell _f)_U}_{L_1(\Dcal)}\right)~,
\]
thus overall
\begin{align*}
\sup_{f\in\Fcal}\left|L_\Dcal(f)-L_S(f)\right|
\leq \alpha &+ \max_{(\ell_f)_L}\left(\norm{(\ell_f)_L}_{L_1(\Dcal)} -\norm{(\ell _f)_L}_{L_1(\Dcal_n)}\right)
\\ &+\max_{(\ell_f)_U}\left(\norm{(\ell_f)_U}_{L_1(\Dcal_n)} -\norm{(\ell _f)_U}_{L_1(\Dcal)}\right)~.
\end{align*}
In order to bound the right-hand side, all that is left is a standard application of Hoeffding's inequality
with a union bound over the finite bracket class, whose size is bounded by $\Nbr(\Fcal,L_1(\mu),\alpha)$
due to \eqref{eq: loss bracket}. Minimizing over $\alpha>0$ completes the proof.

\subsection{Proof of Theorem~\ref{thm: alg}} \label{sec: alg proof}

\begin{algorithm}
\begin{algorithmic}[1] 
\caption{Approximate extension} \label{alg: approx reg}
\STATE \textbf{Input:}
Sample $S=(X_i)_{i=1}^{n}$, labels $(\widehat{f}(X_i))_{i=1}^{n}$, exponent $\beta\in(0,1]$, accuracy parameter $\gamma>0$.
\STATE \textbf{Preprocessing:}
\STATE Sort $(X_1,\dots,X_n)$ according to
\[
w(X_i)=\max_{j\neq i}\frac{\left|\widehat{f}(X_i)-\widehat{f}(X_j)\right|}{\rho(X_i,X_j)^\beta}~.
\]
\STATE Let $S'\subset\{X_1,\dots,X_n\}$ consist of the $n-\lfloor\gamma n\rfloor$ points with smallest $w(X_i)$ value.
\STATE Let $A\subset S'$ be a $\gamma^{1/\beta}$ net of $S'$.
\STATE \textbf{Inference:}
\STATE For any $x\in\Omega$, compute
\[
(u^*,v^*)=\underset{(u,v)\in A\times A}{\arg\max}~\frac{\widehat{f}(v)-\widehat{f}(u)}{\rho(x,u)^\beta+\rho(x,v)^\beta}
\]
and set
\[
f(x):=\widehat{f}(u^*)+\frac{\rho(x,u^*)^\beta}{\rho(x,u^*)^\beta+\rho(x,v^*)^\beta}(\widehat{f}(v^*)-\widehat{f}(u^*))~.
\]
\end{algorithmic}
\end{algorithm}

We will state two propositions due to \citet{hanneke2023near} which, together with Theorem~\ref{thm: gen}, will serve as the main components of the proof.

\begin{proposition}[\citealp{hanneke2023near}] \label{prop: emp smooth}
    Let $f:\Omega\to[0,1]$
    and $\mu$ be any distribution over $\Omega$. 
    Then with probability at least $1-\delta$ over drawing a sample it holds that 
    \[
    \widehat{\Lambda}_{f}^\beta
    \leq 5\log^2(2n/\delta)\overline{\Lambda}_f^\beta(\mu)~.
    \]
\end{proposition}

\begin{proposition}[\citealp{hanneke2023near}] \label{prop: approx extend g}
Algorithm~\ref{alg: approx reg} is an algorithm with $\widetilde{O}(n^2)$ preprocessing time and $O(n^2)$ inference time, that given any $\gamma>0$, a sample $S\sim\Dcal^n$ and any function $\widehat{f}:S\to[0,1]$, provided that $n\geq N$ for 
\[
N=\widetilde{O}\left(\frac{\Ncal_{\Omega}(\gamma)+\log(1/\delta)}{\gamma}\right)~,
\]
constructs a function $f:\Omega\to[0,1]$ such that with probability at least $1-\delta:$
    \begin{itemize}
        \item $L_S(f)\leq L_S(\widehat{f})+\gamma(1+2\widehat{\Lambda}_{\widehat{f}}^\beta)$.
        \item $\overline{\Lambda}^\beta_f(\mu)\leq 5\widehat{\Lambda}_{\widehat{f}}^\beta$~.
    \end{itemize}
\end{proposition}

We are now ready to prove Theorem~\ref{thm: alg}.
Let $\delta'=\frac{\delta}{3}$, and fix $\alpha,\gamma>0$ to be determined later.
Denote $L^*=\inf_{f\in\overline{\Hol}_{H}^\beta(\Omega,\mu)}L_\Dcal(f)$, and let $f^*\in\overline{\Hol}_{H}^\beta(\Omega,\mu)$ be such that $L_\Dcal(f^*)\leq L^*+\alpha$.
We will now describe two desirable events that hold with high probability over drawing the sample $S\sim\Dcal^n$, which we will condition on throughout the rest of the proof.
Consider the event in which
\begin{equation} \label{eq: emp smooth event}
\widehat{\Lambda}_{f^*}^\beta
\leq \widehat{H}:= 5\log^2(2n/\delta')H~,
\end{equation}
and note that this event holds with probability at least $1-\delta'$ according to Proposition~\ref{prop: emp smooth}.
Further consider the event in which for all $f\in\overline{\Hol}_{5\widehat{H}}^\beta(\Omega,\mu):$
\begin{equation} \label{eq: uc event}
\left|L_{\Dcal}(f)-L_S(f)\right|
=\alpha+\widetilde{O}\left(\sqrt{\frac{\Ncal_{\Omega}\left(\left(\frac{\alpha}{640\widehat{H}\log(1/\alpha)}\right)^{1/\beta}\right)+\log(1/\delta')}{n}}\right)~,
\end{equation}
and note that this event holds with probability at least $1-\delta'$ according to Theorem~\ref{thm: gen}.
In particular, since $f^*\in\overline{\Hol}_{H}^\beta(\Omega,\mu)\subset \overline{\Hol}_{5\widehat{H}}^\beta(\Omega,\mu)$, we get that as long as 
\begin{equation} \label{eq: n large}
n=\widetilde{\Omega}\left(\frac{\Ncal_{\Omega}\left(\left(\frac{\alpha}{640\widehat{H}\log(1/\alpha)}\right)^{1/\beta}\right)+\log(1/\delta')}{\alpha^2}\right)~,   
\end{equation}
it holds that
\begin{equation} \label{eq: emp loss}
L_{S}(f^*)\leq L_{\Dcal}(f^*)+2\alpha\leq L^*+3\alpha~.    
\end{equation}
Thus, by solving the following feasible linear program over the variables $\left(\widehat{f}(X_i),\mathrm{err}_i,\widehat{H}_i\right)_{i=1}^{n}:$
\begin{align}
\text{Minimize}~~~~~&\sum_{i=1}^{n}\mathrm{err}_i \nonumber
\\
\text{subject to}~~~~~&\mathrm{err}_i\geq \left|\widehat{f}(X_i)-Y_i\right|~~~~~~~~~~~~~~~~~~~~~~~~~~~~~~~~~~~\forall i\in[n]
\label{eq: err}
\\
&0\leq \widehat{f}(X_i)\leq 1~~~~~~~~~~~~~~~~~~~~~~~~~~~~~~~~~~~~~~~~~~~~\forall i\in[n]
\nonumber
\\
&\frac{1}{n}\sum_{i=1}^{n}\widehat{H}_i\leq 5\widehat{H}
~~~~~~~~~~~~~~~~~~~~~~~~~~~~~~~~~~~~~~~~~~~~\forall i\in[n]
\label{eq: sum Lhat_i}
\\
&|\widehat{f}(X_i)-\widehat{f}(X_j)|\leq \widehat{H}_i\cdot\rho(X_i,X_j)^\beta
~~~~~~~~~~\forall i, j\in[n]:X_i\neq X_j
\label{eq: Lhat_i bound}
\end{align}
it is possible to find $(\widehat{f}(X_1),\dots,\widehat{f}(X_n))$ so that 
\[
L_S(\widehat{f})
=\sum_{i=1}^{n}\left|\widehat{f}(X_i)-Y_i\right|
\overset{\text{\eqref{eq: err}}}{\leq} \sum_{i=1}^{n}\mathrm{err}_i
\overset{(\star)}{\leq} L^*+4\alpha
\]
and 
\[
\widehat{\Lambda}_{\widehat{f}}^\beta 
=\frac{1}{n}\sum_{i=1}^{n}\max_{X_j\neq X_i}\frac{|\widehat{f}(X_i)-\widehat{f}(X_j)|}{\rho(X_i,X_j)^\beta}
\overset{\text{\eqref{eq: Lhat_i bound}}}{\leq} \frac{1}{n}\sum_{i=1}^{n}\widehat{H}_i
\overset{\text{\eqref{eq: sum Lhat_i}}}{\leq} 5\widehat{H}~,
\]
within polynomial time.
Indeed, the feasibility is observed by considering the variable assignment
\begin{align}
\widehat{f}(X_i)&=f^*(X_i) \label{eq: assignment 1}
\\
\mathrm{err}_i&=|f^*(X_i)-Y_i| \label{eq: assignment 2}
\\
\widehat{H}_i&=\max_{X_j \neq X_i}\frac{\left|f^*(X_i)-f^*(X_j)\right|}{\rho(X_i,X_j)^\beta} \label{eq: assignment 3}
~,
\end{align}
since Eqs.~(\ref{eq: assignment 1}) and (\ref{eq: assignment 2}) imply \eqref{eq: err};  Eqs.~(\ref{eq: assignment 1}) and (\ref{eq: assignment 3}) imply \eqref{eq: Lhat_i bound}; and \eqref{eq: emp smooth event} implies \eqref{eq: sum Lhat_i}. Moreover, $(\star)$ follows as long as the program is solved up to accuracy at most $\alpha$ due to \eqref{eq: emp loss}.
The runtime required for solving the program with $O(n^2)$ constraints up to accuracy at most $\alpha$ is bounded, according to the currently best known complexity of linear programming, by $\widetilde{O}\left((n^2)^\omega\right)=\widetilde{O}\left(n^{2\omega}\right)$ where $\omega\approx2.37$ is the current matrix multiplication exponent \citep{cohen2021solving}.

With such $\widehat{f}$ in hand, we can apply Algortithm~\ref{alg: approx reg} in order to obtain $f:\Omega\to[0,1]$, 
whose guaranteed by Proposition~\ref{prop: approx extend g} to satisfy with probability at least $1-\delta':$
\[
L_S(f)
\leq  L_S(\widehat{f})+\gamma(1+2\widehat{\Lambda}_{\widehat{f}}^\beta)
\leq L^* + 4\alpha + \gamma(1+5\widehat{H})
\]
and
\[
\overline{\Lambda}^\beta_f(\mu)\leq5\widehat{\Lambda}_{\widehat{f}}^\beta\leq 5\widehat{H}=25\log^2(2n/\delta')H~.
\]
By \eqref{eq: uc event} and \eqref{eq: n large}, the latter property ensures that
\[
L_\Dcal(f)
\leq
L_S(f)+2\alpha
\leq L^* + 6\alpha + \gamma(1+5\widehat{H})~.
\]
Setting
\[
\alpha=\frac{\eps}{12}~,
~~~\gamma=\frac{\eps}{2+10\widehat{H}}=\widetilde{\Theta}\left(\frac{\eps}{H}\right)~,
\]
and applying the union bound
yields
\[
L_\Dcal(f)\leq L^*+\eps
\]
with probability at least $1-3\delta'=1-\delta$.

\paragraph{Acknowledgments.}

AK is partially supported by the Israel Science Foundation (grant No. 1602/19), an Amazon Research Award, and the Ben-Gurion University Data Science Research Center.
GK is supported by an
Azrieli Foundation graduate fellowship.

\bibliographystyle{plainnat}
\bibliography{bib}

\end{document}